\documentclass[11pt]{article}
\usepackage{eacl2017}
\usepackage{times}
\usepackage{url}
\usepackage{latexsym}
\usepackage{amsmath}
\newcommand{\matr}[1]{\mathbf{#1}}
\usepackage{algorithm}
\usepackage{algpseudocode}
\usepackage{booktabs}
\usepackage{graphicx}
\usepackage{subcaption}
\graphicspath{ {./} }
\usepackage{booktabs}

\usepackage[draft]{hyperref}

\usepackage{amsmath}
\usepackage{centernot}
\usepackage{array,multirow}
\usepackage[font={footnotesize}]{caption}
\usepackage{enumitem}
\usepackage{float}

\eaclfinalcopy % Uncomment this line for the final submission
\algnewcommand\algorithmicforeach{\textbf{for each}}
\algdef{S}[FOR]{ForEach}[1]{\algorithmicforeach\ #1\ \algorithmicdo}

\title{Inducing Regular Grammars Using Recurrent Neural Networks}

\author{
Mor Cohen\thanks{$\quad$ Authors equally contributed to this work.} \\
 Tel-Aviv University \\
  {\small \tt morco@outlook.com} \\\And
 Avi Caciularu\footnotemark[1] \\
 Tel-Aviv University \\
  {\small \tt avi.c33@gmail.com} \\\And
 Idan Rejwan\footnotemark[1] \\
 Tel-Aviv University \\
  {\small \tt idanrejwan91@gmail.com} \\\And
 Jonathan Berant \\
 Tel-Aviv University \\
  {\small \tt joberant@cs.tau.ac.il} \\\mbox{}}
\author{Mor Cohen\thanks{$\quad$ Authors equally contributed to this work.}, Avi Caciularu\footnotemark[1], Idan Rejwan\footnotemark[1], Jonathan Berant \\
\textit{School of Computer Science, }
\textit{Tel-Aviv University, }
Tel-Aviv, 6997801 Israel \\\texttt{morco@outlook.com,avi.c33@gmail.com,} \\ \texttt{idanrejwan@mail.tau.ac.il,joberant@cs.tau.ac.il}
}
\date{}

\begin{document}
\maketitle
\setlength{\abovedisplayskip}{3pt}
\setlength{\belowdisplayskip}{3pt}

\begin{abstract}
Grammar induction is the task of learning a grammar from a set of examples. Recently, neural networks have been shown to be powerful learning machines that can identify patterns in streams of information. In this work we investigate their effectiveness in inducing a regular grammar from data, without any assumptions about the grammar. We train a recurrent neural network to distinguish between strings that are in or outside a regular language, and utilize an algorithm for extracting the learned finite-state automaton. We apply this method to several regular languages and find unexpected results regarding the connections between the network's states that may be regarded as evidence for generalization.

\end{abstract}

\section{Introduction}

Grammar induction is the task of learning a grammar from a set of examples, thus constructing a model that captures the patterns within the observed data. It plays an important role in scientific research of sequential phenomena, such as human language or genetics.
We focus on the most basic level of grammars - regular grammars. That is, the set of all languages that can be decided by a Deterministic Finite Automaton (DFA).

Inducing regular grammars is an old and extensively studied problem \cite{de10}. However, most suggested methods involve prior assumptions about the grammar being learned. In this work, we aim to induce a grammar from examples that are in or outside a language, without any assumption on its structure.

Recently, neural networks were shown to be powerful learning models for identifying patterns in data, including language-related tasks \cite{li16,kuncoro1lingpeng}. This work investigates how good neural networks are at inducing a regular grammar from data. More specifically, we investigate whether RNNs, a neural network that specializes in processing sequential streams, can learn a DFA from data.

RNNs are suitable for this task since they resemble DFAs. At each time step the network has a current state, and given the next input symbol it produces the next state. Formally, let $s_{t}$ be the current state and $x_{t+1}$ the next input symbol, then the RNN computes the next state $s_{t+1} = \delta(s_{t},x_{t+1})$, where $\delta$ is the function learned by the RNN. Consequently, $\delta$ is actually a transition function between states, similar to a DFA.

This analogy between RNNs and DFAs suggests a way to understand RNNs. It enables us to "open the black box" and analyze the network by converting it into the corresponding DFA and examining the learned language. 

Inspired by that, we explore a method for grammar induction. Given a labeled dataset of strings that are in and outside a language, we wish to train a network to classify them. If the network succeeds, it must have learned the latent patterns underlying the data.
This allows us to extract the states used by the network and reconstruct the grammar it had learned.

There is one major difference between the states of DFAs and RNNs. While the former are discrete and finite, the latter are continuous. In theory, this difference makes RNNs much more powerful than DFAs \cite{si95}. However in practice, simple RNNs
\footnote{Without the aid of additional memory such as in LSTMs \cite{hochreiter1997long}}
are not strong enough to deal with languages beyond the regular domain \cite{ge01}.

Similar ideas have already been investigated in the 1990s \cite{cleeremans1989finite,giles1990higher,elman1991distributed,omlin1996extraction,morris1998connectionist,tino1998finite}\footnote{For a more thorough survey of relevant papers, see section 5.13 in \cite{schmidhuber2015deep}}. These works also presented techniques to extract a DFA from a trained RNN. However, most of them included a priori quantization of the RNN’s continuous state space, yielding exponential number of states even for simple grammars. Instead, several works used clustering techniques for quantizing the state space, which yielded much smaller number of states \cite{zeng1993learning}. However, they fixed the number of clusters in advance. In this work, we introduce a novel technique for extracting a DFA from an RNN using clustering without the need to know the number of cluster. For that purpose, we present a heuristic to find the most suitable number of states for the DFA, making the process much more general and unconstrained.

\section{Problem Statement}

Given a regular language $\mathcal{L}$ and a labeled dataset $\left\{(X_i,y_i)\right\}$ of strings $X_i$ with binary labels $y_i = 1 \Leftrightarrow X_i \in \mathcal{L}$, 
the goal is to output a DFA $A$ such that $A$ accepts $\mathcal{L}$, i.e., $L(A)=\mathcal{L}$.

Since the target language $\mathcal{L}$ is unknown, we relax this goal to $A(\bar{X}_i) = \bar{y}_i$. Namely, $A$ accepts or rejects correctly on a test set $\{(\bar{X}_i,\bar{y}_i)\}$ that has not been used for training. %\\ 
Accordingly, the accuracy of $A$ is defined as the proportion of strings classified correctly by $A$.

\section{Method}
The method consists of the following main steps.
\footnote{\scriptsize Python code is available at \url{https://github.com/acrola/RnnInduceRegularGrammar}}
\begin{enumerate}[topsep=0pt,itemsep=0pt,parsep=0pt]
  \item \textbf{Data Generation} - creating a labeled dataset of positive and negative strings: $15,000$ strings for training the network, a validation set of $10,000$ strings for constructing the DFA, and a test set of $10,000$ strings for evaluating the results. 
  \item \textbf{Learning} - training an RNN within 15 epochs to classify the dataset with high accuracy- in almost all the conducted experiments (fully described in the next section), a nearly perfect validation accuracy is achieved, approximately $99\%$.
  \item \textbf{DFA Construction} - extracting the states produced by the RNN, quantizing them and constructing a minimized DFA.
\end{enumerate}

\subsection{Data Generation}
The following method was used to create a balanced dataset of positive and negative strings. Given a regular expression we randomly generate sequences out of it. As for the negative strings, two different methods were used. The first method was to generate random sequences of words from the vocabulary, such that their length distribution is identical to the positive ones. The other approach was to generate ungrammatical strings which are almost identical to the grammatical ones, making the learning more difficult for the RNN. We generated the negative strings by applying minor modifications such as word deletions, additions or movements. Both methods did not yield any significant difference in the results, thus we show only results for the first method.

\subsection{Learning}
We used the most basic architecture of RNNs, with one layer and cross entropy loss.
In more detail, the RNN's transition function is a single fully-connected layer given by 
\[
s_{t}=\tanh\left(\matr{W}s_{t-1}+\matr{U}x_{t}+v\right).
\]
Prediction is made by another fully-connected layer which gets the RNN's final state $s_n$ as an input and returns a prediction $\hat{y}\in [0,1]$ by, 
\[
\hat{y} = \sigma(\matr{A}\sigma(\matr{B}s_{n}+c)+d).
\]
where $\sigma$ stands for the sigmoid function, $\matr{W},\matr{U},\matr{A},\matr{B}$ are learned matrices and $v,c,d$ are learned vectors.\\
The loss used for training is cross entropy,
\[
l = -\frac{1}{n} \sum_{i=1}^n y_i \log \hat{y}_i 
+ \left(1 - y_i\right) \log \left(1 - \hat{y}_i\right),
\]
where $y_1,\dots,y_n$ are the true labels and $\hat{y}_1,\dots,\hat{y}_n$ are the network's predictions. To optimize our loss, we employ the Adam algorithm \cite{kingma2014adam}

Our goal is to reach perfect accuracy on the validation set, in order to make sure the network succeeded in generalizing and inducing the regular grammar underlying the dataset. This assures that the DFA we extract later is reliable as much as possible. 

\subsection{DFA Construction}
When training is finished, we extract the DFA learned by the network. This process consists of the following four steps.

\paragraph{Collecting the states}
First, we collect the RNN's continuous state vectors by feeding the network with strings from the validation set and collecting the states it outputs while reading them.

\paragraph{Quantization}
After collecting the continuous states, we need to transform them into a discrete set of states. We can achieve this by simply using any conventional clustering method with the Euclidean distance measure. 
More specifically, we use the $K$-means clustering algorithm, where $K$ is taken to be the minimal value such that the quantized graph's classifications match the network's ones with high rate.
That is, for each $K$ we build the quantized DFA (as we describe later) and count the number of matches between the DFA's classifications and the network's over a validation set. We return the minimal $K$ that exceeds 99\% matches.
It should be noted that the initial state is left as is and is not associated with any of the clusters.

\paragraph{Building the DFA}
Given the RNN's transition function $\delta$ and the clustering method $c$, we use the following algorithm to build the DFA.
\begin{algorithm}[H]
\centering
\caption{DFA Construction}
\begin{algorithmic}
\State $V,E\leftarrow \phi, \phi$
\ForEach {sequence $X_i$}
\State $s_{t-1}, v_{t-1}\leftarrow s_0, c(s_0)$
\ForEach {symbol $x_{t+1}\in X_i$}
\State $s_{t}\leftarrow \delta(s_{t-1},x_{t})$
\State $v_{t}\leftarrow c(s_{t})$
\State Add $v_{t}$ to $V$
\State Add $(v_{t-1},x_t)\rightarrow v_{t}$ to $E$
\State $s_{t-1}, v_{t-1} \leftarrow s_{t}, v_{t} $
\EndFor
\State Mark $v_{t}$ as accept if $\hat{y}_i=1$
\EndFor
\State \textbf{return} $V,E$
\end{algorithmic}
\end{algorithm}

Finally, we use the Myhill-Nerode algorithm in order to find the minimal equivalent DFA \cite{downey2012parameterized}.

\section{Experiments}
To demonstrate our method, we applied it on the following few regular expressions.
\subsection{Simple Binary Regexes}
\label{subsec:binregex}
\begin{figure}
        \begin{subfigure}[b]{0.2\textwidth}
                \centering
                \includegraphics[width=.85\linewidth]{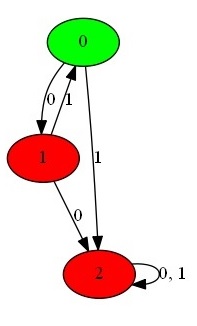}
                \caption{$(01)*$}
                \label{fig:regexa}
        \end{subfigure}%
        \begin{subfigure}[b]{0.2\textwidth}
                \centering
                \includegraphics[width=.85\linewidth]{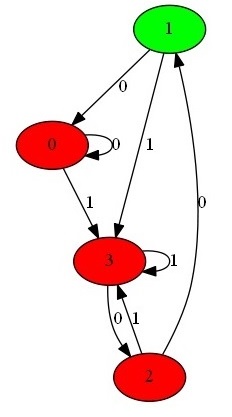}
                \caption{$\text{(0}|\text{1)}^*\text{100}$}
                \label{fig:regexb}
        \end{subfigure}
        \caption{Minimized DFAs for binary regexes}\label{fig:minidfas}
\end{figure}

The resulting DFAs for the two binary regexes $(01)*$ and $(0|1)*100$ are shown in Figure \ref{fig:minidfas}.

It can be observed that the method produced perfect DFAs that accept exactly the given languages. The DFAs accuracy was indeed 100\%.

A finding worth mentioning is the emergence of cycles within the continuous states transitions. That is, the RNN before quantization mapped new states into the exact same state it has already seen before. This is surprising because if we think of the continuous states as random vectors, the probability to see an exact vector twice is zero. This finding, which was reproduced for several regexes, may be an evidence for generalization as we discuss later. 

\subsection{Part of Speech Regex}
\begin{figure}
\centering
\includegraphics[width=.25\linewidth]{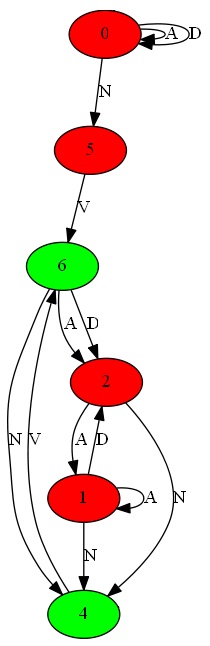}
   \caption{Minimized DFA for synthetic POS regex}
\label{fig:regexc}
\end{figure}

To test our model on a more complicated grammar, we created a synthetic regex inspired by natural language. The regex we used describes a simplified part-of-speech grammar,
\[
\textsc{Det? Adj}* \textsc{Noun Verb (Det? Adj}* \textsc{Noun)?}
\] 
The resulting DFA is shown in Figure \ref{fig:regexc}.

The DFA's accuracy was 99.6\%, i.e. the learned language is not exactly the target one, but is very close. For example, it accepts sentences like
\[
\textbf{The nice boy kissed a beautiful lovely girl}
\]
and rejects sentences like
\[
\textbf{The boy nice}
\]
However, by examining the DFA we can find sentences that the network misclassifies. For example, it accepts sentences like
\[
\textbf{The the boy stands}
\]

Inspecting the DFA's errors might be meaningful also for training, as a technique for targeted data augmentation. By the pumping lemma, each of the states where the network is wrong stands for an infinite class of sequences that end at the same state. In other words, those states are actually a "formula" for generating as much data as we want, such that the network is wrong. This way, the network can be re-trained on its own errors. 

\section{Discussion}

\subsubsection*{Learning}
The network reached 100\% accuracy quickly on the synthetic datasets. This may indicate that deciding a regular language is a reasonable task for an RNN, and illustrates the similarity between RNNs and DFAs discussed earlier.

\subsubsection*{Emergence of cycles}
The emergence of cycles within the continuous states, mentioned in \hyperref[subsec:binregex]{experiment 4.1}
, might be understood as an evidence for generalization. Having learned those cycles means that for an infinite set of sequences the network will always traverse the same path and predict the same label. In other words, the model has generalized for sequences of arbitrary length \emph{before} quantization. 

It should be noted that such cycles have also been noticed in \cite{tino1998finite}. Nevertheless, in their work the learning objective was predicting the next state rather than classifying the whole sentence. As a result, the emergence of cycles is expected, since the network was forced to learn them by the supervision. This differs from the case of classification, as learning the states and the cycles is not supervised.

\subsubsection*{Quantization}
The process used for quantizing the states may introduce conflicts, if two different states in the same cluster lead to two different clusters for the same input. 
However, all of our experiments did not yield any conflict. This means that the clusters do reflect well the RNN's different states.

This is reasonable due to the continuity of the RNN's transition function, which maps "close" states into "close" states in terms of Euclidean distance.
Thus, states within the same cluster will have similar transitions and therefore be mapped into the same cluster.

\begin{figure}[H]
\centering
\includegraphics[width=.70\linewidth]{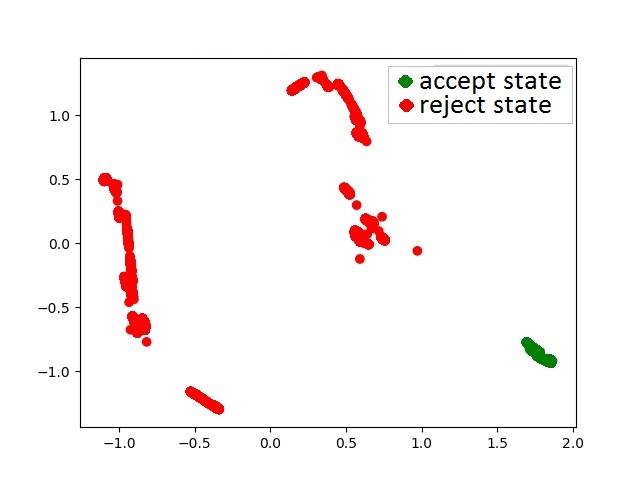}
   \caption{RNN's continuous states}
\label{fig:quant}
\end{figure}

Another way to confirm the clusters validity is to check their compatibility with the network's decisions, i.e., whether accepting or rejecting states are clustered together. Figre \ref{fig:quant} presents the continuous vectors for the binary regex $(0|1)*100$\footnote{To reduce the vectors dimensions we used PCA.}. Clearly, the states are divided into five distinct clusters and only one of them is accepting.

\section{Conclusions}
We investigated a method for grammar induction using recurrent neural networks. This method gives some insights about RNNs as a learning model, and raises several questions, for example regarding the cycles within the continuous states.

Quantization via clustering proved itself to reflect well the true states of the network. Identifying an infinite set of vectors as one state may reduce the network's sensitivity to noise. Thus, states quantization during or after training may be considered as a technique for injecting some robustness to the model and reducing overfitting.

Finally, this method may serve as a tool for scientific research, by finding regular patterns within a real-world sequence. For example, it would be interesting to use this method for natural language, more specifically to induce grammar rules of phonology, which is claimed to be regular \cite{ka94}. Another example is to use it to find regularity within the structure of DNA, which is also regarded as regular \cite{gusfield1997algorithms}.

\section{Acknowledgements}
This work was supported by the Yandex Initiative in Machine Learning.

\bibliography{eacl2017}

\begin{thebibliography}{}

\bibitem[\protect\citename{Cleeremans \bgroup et al.\egroup
  }1989]{cleeremans1989finite}
Axel Cleeremans, David Servan-Schreiber, and James~L McClelland.
\newblock 1989.
\newblock Finite state automata and simple recurrent networks.
\newblock {\em Neural computation}, 1(3):372--381.

\bibitem[\protect\citename{De~la Higuera}2010]{de10}
Colin De~la Higuera.
\newblock 2010.
\newblock {\em Grammatical inference: learning automata and grammars}.
\newblock Cambridge University Press.

\bibitem[\protect\citename{Downey and Fellows}2012]{downey2012parameterized}
Rodney~G Downey and Michael~Ralph Fellows.
\newblock 2012.
\newblock {\em Parameterized complexity}.
\newblock Springer Science \& Business Media.

\bibitem[\protect\citename{Elman}1991]{elman1991distributed}
Jeffrey~L Elman.
\newblock 1991.
\newblock Distributed representations, simple recurrent networks, and
  grammatical structure.
\newblock {\em Machine learning}, 7(2-3):195--225.

\bibitem[\protect\citename{Gers and Schmidhuber}2001]{ge01}
Felix~A Gers and E~Schmidhuber.
\newblock 2001.
\newblock Lstm recurrent networks learn simple context-free and
  context-sensitive languages.
\newblock {\em IEEE Transactions on Neural Networks}, 12(6):1333--1340.

\bibitem[\protect\citename{Giles \bgroup et al.\egroup }1990]{giles1990higher}
C~Lee Giles, Guo-Zheng Sun, Hsing-Hen Chen, Yee-Chun Lee, and Dong Chen.
\newblock 1990.
\newblock Higher order recurrent networks and grammatical inference.
\newblock In {\em Advances in neural information processing systems}, pages
  380--387.

\bibitem[\protect\citename{Gusfield}1997]{gusfield1997algorithms}
Dan Gusfield.
\newblock 1997.
\newblock {\em Algorithms on strings, trees and sequences: computer science and
  computational biology}.
\newblock Cambridge university press.

\bibitem[\protect\citename{Hochreiter and Schmidhuber}1997]{hochreiter1997long}
Sepp Hochreiter and J{\"u}rgen Schmidhuber.
\newblock 1997.
\newblock Long short-term memory.
\newblock {\em Neural computation}, 9(8):1735--1780.

\bibitem[\protect\citename{Kaplan and Kay}1994]{ka94}
Ronald~M Kaplan and Martin Kay.
\newblock 1994.
\newblock Regular models of phonological rule systems.
\newblock {\em Computational linguistics}, 20(3):331--378.

\bibitem[\protect\citename{Kingma and Ba}2014]{kingma2014adam}
D.~Kingma and J.~Ba.
\newblock 2014.
\newblock Adam: A method for stochastic optimization.
\newblock {\em arXiv preprint arXiv:1412.6980}.

\bibitem[\protect\citename{Kuncoro and Ballesteros}]{kuncoro1lingpeng}
Adhiguna Kuncoro and Miguel Ballesteros.
\newblock Lingpeng kong, chris dyer, graham neubig, and noah a. smith. 2017.
  what do recurrent neural network grammars learn about syntax.
\newblock In {\em Proceedings of the 15th Conference of the European Chapter of
  the Association for Computational Linguistics}, volume~1, pages 1249--1258.

\bibitem[\protect\citename{Linzen \bgroup et al.\egroup }2016]{li16}
Tal Linzen, Emmanuel Dupoux, and Yoav Goldberg.
\newblock 2016.
\newblock Assessing the ability of lstms to learn syntax-sensitive
  dependencies.
\newblock {\em arXiv preprint arXiv:1611.01368}.

\bibitem[\protect\citename{Morris \bgroup et al.\egroup
  }1998]{morris1998connectionist}
William~C Morris, Garrison~W Cottrell, and Jeffrey Elman.
\newblock 1998.
\newblock A connectionist simulation of the empirical acquisition of
  grammatical relations.
\newblock In {\em International Workshop on Hybrid Neural Systems}, pages
  175--193. Springer.

\bibitem[\protect\citename{Omlin and Giles}1996]{omlin1996extraction}
Christian~W Omlin and C~Lee Giles.
\newblock 1996.
\newblock Extraction of rules from discrete-time recurrent neural networks.
\newblock {\em Neural networks}, 9(1):41--52.

\bibitem[\protect\citename{Schmidhuber}2015]{schmidhuber2015deep}
J{\"u}rgen Schmidhuber.
\newblock 2015.
\newblock Deep learning in neural networks: An overview.
\newblock {\em Neural networks}, 61:85--117.

\bibitem[\protect\citename{Siegelmann and Sontag}1995]{si95}
Hava~T Siegelmann and Eduardo~D Sontag.
\newblock 1995.
\newblock On the computational power of neural nets.
\newblock {\em Journal of computer and system sciences}, 50(1):132--150.

\bibitem[\protect\citename{Tino \bgroup et al.\egroup }1998]{tino1998finite}
Peter Tino, Bill~G Horne, and C~Lee Giles.
\newblock 1998.
\newblock Finite state machines and recurrent neural networks--automata and
  dynamical systems approaches.
\newblock Technical report.

\bibitem[\protect\citename{Zeng \bgroup et al.\egroup }1993]{zeng1993learning}
Zheng Zeng, Rodney~M Goodman, and Padhraic Smyth.
\newblock 1993.
\newblock Learning finite state machines with self-clustering recurrent
  networks.
\newblock {\em Neural Computation}, 5(6):976--990.

\end{thebibliography}
\bibliographystyle{eacl2017}

\end{document}